%% file: iclr2027_conference.tex
\documentclass{article} 
\usepackage{iclr2027_conference,times}

\input{math_commands.tex}

\usepackage{hyperref}
\usepackage{url}
\usepackage{physics}
\usepackage{wrapfig}
\usepackage{subcaption}
\usepackage{xspace}
\usepackage{booktabs}       
\usepackage{graphicx}
\usepackage{multirow}
\usepackage{algorithm}
\usepackage{algpseudocode}

\DeclareRobustCommand{\model}{Seq-Flow\xspace}

\title{Seq-Flow: Efficient Probabilistic Forecasting with Self-Rollout Error Control}

\author{Yinan Huang\\
Georgia Institute of Technology\\
\texttt{yhuang903@gatech.edu} \\
\And 
Shitij Govil \\ 
Georgia Institute of Technology\\
\texttt{sgovil9@gatech.edu} \\
\And
Bo Dai\\
Georgia Institute of Technology\\
\texttt{bodai@cc.gatech.edu} \And 
Pan Li \\
Georgia Institute of Technology\\
\texttt{panli@gatech.edu} 
}

\iclrfinalcopy 
\begin{document}


\maketitle

\begin{abstract}

Many scientific forecasting tasks require updating a distribution over future trajectories as new observations arrive. Conventional diffusion and flow models generate each forecast from Gaussian noise, often at the cost of many sampling steps. Warm-start methods reuse earlier predictions to reduce this cost, but their models are not trained to perform the forecast update itself, which can compromise quality under few-step sampling. In this work, we introduce \model, a conditional flow model whose ODE transports samples from the previous forecast distribution to the updated one. Because successive forecasts often differ only modestly, this transport starts from an informative distribution and can produce accurate updates with few flow evaluations. Recursive reuse also creates a challenge: errors in one forecast become errors in the initial states of subsequent flows. We address this with self-rollout training, in which a moving average copy of the model generates forecasts that initialize later training updates. Unlike self-forcing methods, which reuse generated outputs as conditioning context, \model reuses them as the source of the next flow. Experiments On particle-accelerator beam spill forecasting show \model reduces CRPS by $65\%$ under a few-NFE sampling budget, while remaining competitive with strong baselines on fluid-dynamics forecasting tasks. Although trained on self-rollouts of at most four updates, \model remains accurate over more than 400 consecutive updates. Our code is available at \url{https://github.com/Graph-COM/Seq-Flow}.
\end{abstract}

\input{1_intro}
\input{3_method}

\input{4_related}

\input{5_exp}

\section{Conclusion and Limitations}
\label{sec:conclusion}
We introduce \model, a flow matching approach that tracks online forecast distributions by transporting the previous forecast to the updated one conditioned on new observations, enabling accurate forecasting with few flow evaluations. To control the source-distribution error induced by recursive forecast reuse, we proposed self-rollout training with EMA-based rollout generation. Across several scientific forecasting tasks, \model achieves strong few-step performance and remains stable over long recursive deployments. A limitation of our approach is that self-rollout training introduces extra training cost.







\subsubsection*{Acknowledgments}
We would like to thank Hans Hao-Hsun Hsu for his insightful discussions on this work. 

This work is primarily supported by National Science Foundation (NSF) under awards PHY-2117997, IIS-2239565, IIS-2428777, and CCF-2402816; the Google Cloud Research Credit, 2026; the Nvidia Academic Award, 2026. Prof. Bo Dai would like to acknowledge support from NSF ECCS-2401391, NSF IIS-2403240, NSF IIS-2552007 and ONR N000142512173.

\bibliography{iclr2027_conference}
\bibliographystyle{iclr2027_conference}

\appendix
\include{6_appendix}

\end{document}

%% file: math_commands.tex
\usepackage{amsmath,amsfonts,bm}

\def\eqref#1{equation~\ref{#1}}

\def\1{\bm{1}}

\DeclareMathAlphabet{\mathsfit}{\encodingdefault}{\sfdefault}{m}{sl}
\SetMathAlphabet{\mathsfit}{bold}{\encodingdefault}{\sfdefault}{bx}{n}



%% file: 1_intro.tex
\section{Introduction}

Faithfully and efficiently modeling the evolving distribution of future trajectories is crucial for scientific forecasting, monitoring, and control. In applications such as weather forecasting~\citep{kalnay2003atmospheric,lorenz2017deterministic}, fluid dynamics~\citep{bewley2001flow}, and magnetic plasma control~\citep{degrave2022magnetic}, noisy observations and partial observability make the future inherently uncertain, requiring models to continually update this distribution as new observations arrive. Canonical filtering methods, such as the Kalman filter and its nonlinear extensions~\citep{kalman1960new,julier1997new}, recursively propagate predictive distributions but require a specified dynamical model, limiting their ability to capture complex and multimodal futures. Recurrent neural models~\citep{shi2015convolutional,chung2015recurrent,salinas2020deepar} learn more expressive dynamics directly from data, but often produce point forecasts or rely on restrictive parametric output distributions, limiting the predictive uncertainty they can represent.




Diffusion models~\citep{sohl2015deep,song2019generative,ho2020denoising} and flow matching~\citep{lipman2023flow,liu2023flow} provide expressive ways to model complex, multimodal distributions, motivating their growing use in forecasting, planning, and other sequential tasks~\citep{chen2024diffusion,gao2023prediff,janner2022planning,wei2024diffphycon}. In conventional approaches, however, each forecast is generated anew from an uninformative Gaussian source. Reaching a high-fidelity sample from this source can require many denoising or ODE solver steps, while using only a few function evaluations often degrades sample quality~\citep{salimans2022progressive,liu2023flow}. In rapidly evolving systems, this quality–compute trade-off directly limits how frequently predictive distributions can be updated as new observations arrive.
To reduce the cost of repeatedly sampling a predictive distribution from Gaussian noise, we propose \model, a flow-matching approach that updates predictive samples as new observations arrive. Unlike a deterministic point forecaster, \model maintains multiple samples representing plausible future trajectories. At each observation time, it uses samples from the previous predictive distribution as the source and transports them toward the updated predictive distribution conditioned on the newly observed history. Because these samples already reflect past observations, they offer an informative starting point that may allow accurate updates with fewer flow evaluations than sampling from Gaussian noise anew. This recursive transport parallels classical filtering, which updates predictive beliefs as observations arrive~\citep{kalman1960new,gordon1993novel}.

However, recursively reusing forecasts creates a train--test shift in the flow's source distribution, a failure mode absent when each generation starts from a fixed Gaussian prior. If training uses ground-truth trajectories as sources, the flow is not exposed to the errors produced by its own recursive updates. At inference, however, each output feeds into the next flow, so these errors can propagate despite conditioning on new observations. This mechanism differs from the autoregressive exposure bias studied in prior work~\citep{bengio2015scheduled,huang2026self}: autoregressive models feed generated outputs into the conditioning history, whereas \model uses generated forecasts to construct the source distribution of the next flow. This source-level feedback is particularly risky under few-step sampling. To address this mismatch, we train \model through self-rollouts generated by an exponential moving average (EMA) copy of the update model. Each generated forecast is used to construct the source for the next training update, allowing the flow to learn from imperfect forecasts of the kind encountered during recursive deployment.

We evaluate \model on particle-accelerator Beam Spill forecasting
~\citep{whitbeck2025fast}, simulated Burgers' dynamics
~\citep{hwang2022solving}, and real-world fluid--structure interaction data
from RealPDEBench~\citep{hu2026realpdebench}. On Beam Spill, \model reduces
CRPS by 65\% relative to a standard diffusion baseline under a matched 3-NFE
sampling budget, while remaining competitive with strong baselines on both
fluid-dynamics tasks. We further show that \model remains stable over long forecast runs: despite training on self-rollouts of at most four updates, it maintains forecast accuracy through more than 400 consecutive updates without substantial error accumulation.

%% file: 3_method.tex
\section{Methodology}

\subsection{Problem Formulation and Flow Matching Preliminary}
\textbf{Problem Formulation.} We consider an online probabilistic forecasting problem for a dynamical system evolving over time $t=1,2,\ldots,T$. Let $x_t$ denote the system state at time $t$, and let $x_{\leq t}:=x_{1:t}$ denote the observed state history up to time $t$. At each time step, the goal is to characterize the distribution of the future $H$-step trajectory $x_{t+1:t+H}$ conditioned on the current history $x_{\leq t}$, where $H$ denotes the prediction horizon. We use $\mathbf{X}_{t}:=x_{t+1:t+H}$ to denote the $H$-step future trajectory. We therefore formulate online probabilistic forecasting as tracking the time-evolving forecasting distribution $p_t := p(\mathbf{X}_{t}| x_{\leq t})$ as the system evolves and new observations become available.

\textbf{Flow Matching.} Flow matching learns a velocity field $v_\theta$ whose ODE $\frac{d}{d\tau} \mathbf{Z}(\tau) = v_\theta(\mathbf{Z}(\tau), \tau)$ transports samples from a source distribution $\mathbf{Z}(0) \sim p_{\mathrm{source}}$ to a target distribution $\mathbf{Z}(1) \sim p_{\mathrm{target}}$ ~\citep{lipman2023flow, liu2023flow}. Using the linear interpolation $\mathbf{Z}(\tau) = (1-\tau)\mathbf{Z}(0) + \tau\mathbf{Z}(1)$, the velocity field is trained by regressing to the conditional displacement $\mathbf{Z}(1)-\mathbf{Z}(0)$, with $\tau \sim \mathrm{Uniform}(0,1)$. At inference, samples from the target distribution are obtained by integrating the learned ODE starting from the source distribution. Throughout the paper, we use $\tau$ to denote flow time and $t$ to denote the physical time of the dynamical system. In our forecasting setting, $\mathbf{X}_t$ denotes the future trajectory at physical time $t$, whereas $\mathbf{Z}_t(\tau)$ denotes the auxiliary variable evolving along the flow used to generate a forecast sample of $\mathbf{X}_t$.

\subsection{Existing Practices for Online Forecasting with Flow Models}
\label{sec:existing}

A common approach to probabilistic forecasting with diffusion or flow models is to formulate each forecast as a conditional generation problem~\citep{rasul2021autoregressive,kollovieh2025flow}. Specifically, at physical time $t$, the model transports samples from a fixed source distribution, typically Gaussian noise, to the predictive distribution
$p_t = p(\mathbf{X}_t|x_{\le t})$, conditioned on the observed history $x_{\le t}$. This gives the flow ODE
\begin{equation}
\label{eq:ode_Gaussian}
\frac{d}{d\tau} \mathbf{Z}_t(\tau)
=
v_\theta\!\left(\mathbf{Z}_t(\tau), \tau; x_{\le t}\right),
\qquad
\mathbf{Z}_t(0) \sim \mathcal{N}(0,I), \quad
\mathbf{Z}_t(1) \sim p_t,
\end{equation}
where $\tau \in [0,1]$ denotes the flow time and $t$ denotes the physical time step.

Equation~\ref{eq:ode_Gaussian} requires repeatedly transporting samples from Gaussian noise to $p_t$. Such noise-to-target transport typically requires multiple neural function evaluations (NFEs), while reducing the integration steps can degrade sample quality~\citep{salimans2022progressive}. It motivates recent work on few-step flow models via flow map distillation~\citep{song2023consistency,geng2025mean}. In contrast, we aim to exploit the temporal continuity of online forecasting to reduce the transport burden itself rather than distillating the flow sampling path.


\subsection{\model: Aligning Flow Transport with Forecast Update}
\label{sec:seqflow}

\begin{figure}[t]
    \centering
    \includegraphics[width=\linewidth]{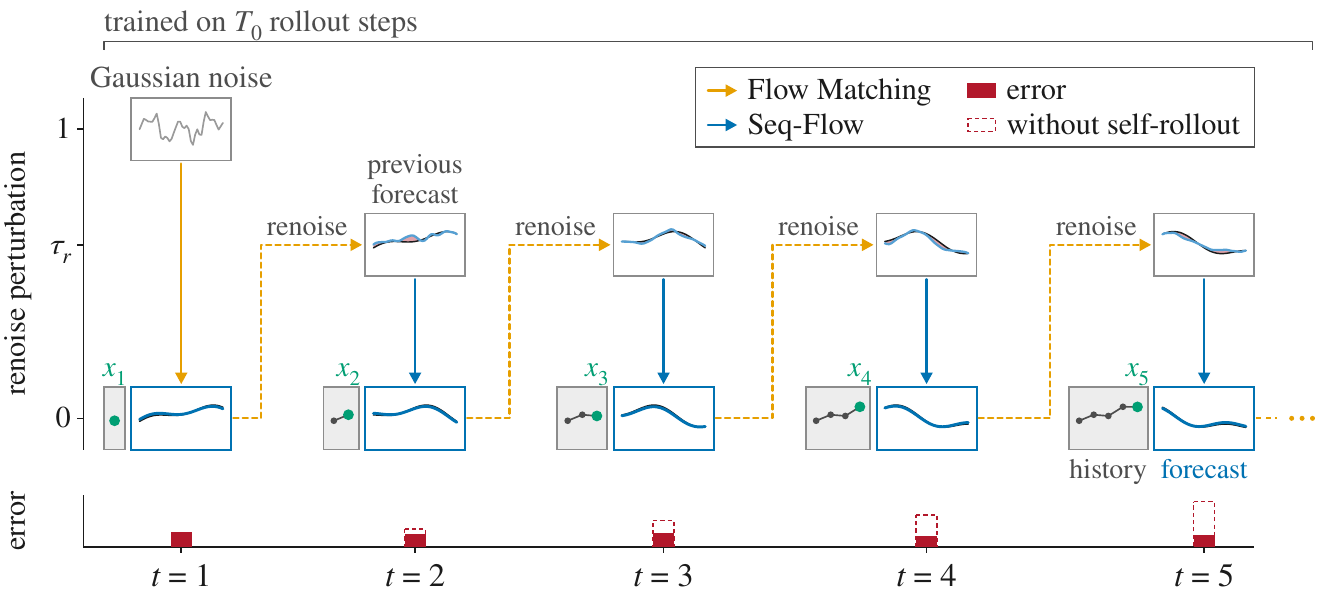}
    \caption{\textbf{\model with self-rollout training.} \model tracks the forecast distribution by modeling the forecast update. It uses a renoised sample from the previous forecast as the source distribution for the next flow, enabling efficient few-step updates. Self-rollout training exposes the model to its own forecast sources, teaching it to correct errors that would otherwise compound across updates.}
    \label{fig:model}
\end{figure}

Our key idea is to formulate the distributional change in the forecasting update itself as the flow transport problem. We propose \model, which aligns the flow transport with the temporal evolution of the forecasting distribution. Instead of transporting samples from a fixed Gaussian source to \(p_t\) at every physical time step, \model directly transports the previous forecasting distribution \(p_{t-1}\) to the updated distribution \(p_t\), via the following flow ODE:
\begin{equation}
\label{eq:ode_align}
\frac{d}{d\tau} \mathbf{Z}_t(\tau)
=
v_\theta(\mathbf{Z}_t(\tau), \tau; x_{\le t}),
\qquad
\mathbf{Z}_t(0)\sim p_{t-1},
\quad
\mathbf{Z}_t(1)\sim p_t.
\end{equation}
Thus, \model learns an update operator between successive forecast distributions by incorporating new observed history. Recursively applying these transports tracks the evolution \(p_1\to p_2\to\cdots\). At the initial step \(t=1\), we define \(p_0:=\mathcal{N}(0,I)\) and generate the first forecast from Gaussian noise. Figure~\ref{fig:model} illustrates the recursive forecast update process of \model. 

\textbf{Temporal continuity enables efficient sampling.}
For temporally smooth dynamical systems, successive predictive distributions
$p_{t-1}$ and $p_t$ typically share substantial structure. Using samples from
$p_{t-1}$ as the source therefore gives the flow a more informative starting
distribution than Gaussian noise, so each update only needs to model the
change in the forecast distribution rather than generating from scratch.
Prior work has shown that replacing an uninformed Gaussian source with a more informative prior can improve generation under limited sampling budgets ~\citep{ren2025prior,scholz2025warm,park2024leveraging}. Instead, Seq-Flow obtains an informative source from the sequential forecasting process itself, using the previous predictive distribution and directly learning the recursive transport $p_{t-1}\to p_t$.

\textbf{Comparison to physical-time flow formulations.}
Recent approaches such as Streaming Flow~\citep{jiang2025streaming} and
ODEWorld~\citep{liu2026odeworld} also align flow ODE dynamics with
physical-time evolution, but in a different manner. First, advancing along
their flow ODE step corresponds to advancing the system in one physical
time step. In contrast, \model uses an entire flow transport integration to model one
physical-time forecast update $p_{t-1}\rightarrow p_t$, allowing multiple
flow evaluations for each update. Second, these methods model an open-loop
temporal evolution under a fixed observation context. Newly arriving
observations are incorporated by initiating or replanning a subsequent
rollout. Seq-Flow instead tracks the evolving forecast distribution in a
closed-loop manner, where each newly observed state directly induces the next
transport.


\subsection{Controlling Recursive Source-distribution Error by Self-Rollout}
\label{sec:self_rollout}

Leveraging the flow ODE to model the forecast update $p_{t-1}\to p_t$ enables efficient sampling but carries sampling errors directly into subsequent updates. At inference, \model reuses its previous
generated forecast $\hat{p}_{t-1}$ as the source distribution of the next flow.
Errors in one forecast therefore directly perturb the initialization of the
subsequent transport. If the model is trained only with ground-truth sources
$p_{t-1}$, it is never exposed to these model-induced source deviations, which
can compound over successive forecast updates.

This error accumulation problem differs from exposure bias~\citep{bengio2015scheduled, huang2026self} in autoregressive
noise-to-target generation. There, each prediction is generated from a fixed
noise source and past errors enter through the conditioning history. In
our setting, the observed history $x_{\le t}$ remains unchanged, while the
model-generated forecast $\hat{\mathbf{X}}_{t-1}$ initializes the next flow, resulting train--test mismatch in the flow source
distribution. We next develop training strategies to mitigate this recursive
source-distribution error.  

\textbf{Self-rollout training with robust source distributions.}
We propose to explicitly unroll the flow model to follow the inference process of forecast update during training. To our knowledge, this is the first
training framework that uses a flow model's own recursive forecast rollouts as
\emph{source samples} for subsequent flow updates. Given a ground-truth trajectory $x_{1:T}^*$ and starting from
$p_0=\mathcal{N}(0,I)$, we recursively generate model-induced forecasts
$p_0 \rightarrow \hat p_1 \rightarrow \cdots \rightarrow \hat p_{T_0}$ using
\eqref{eq:ode_align} (the unrolling step $T_0$ can be far less than the task episode, as we will discuss in Section~\ref{sec:stable_train}). At each step $t$, a generated forecast
$\hat{\mathbf{X}}_{t-1}\sim\hat p_{t-1}$ provides the source for the next
forecast update. To improve robustness to deviations in these recursively
generated sources, we partially renoise each rollout sample as
$
\tilde{\mathbf{X}}_{t-1}=(1-\tau_{r})\hat{\mathbf{X}}_{t-1}
+
\tau_{r}\cdot\mathcal{N}(0,I),
$ where $\tau_{r}\in [0, 1]$ is a renoise level hyperparameter.
After collecting the rollouts $\{\tilde{\mathbf{X}}_t\}_{t=1}^{T_0}$, the model is then trained to transport the resulting source
$\tilde{\mathbf{X}}_{t-1}$ to the ground-truth future trajectory
$\mathbf{X}^*_t=x^*_{t+1:t+H}$ using the flow-matching objective
\begin{equation}
\mathcal{L}(\theta)
=
\sum_t
\mathbb{E}_{\tau\sim\mathrm{Uniform}(0,1)}
\left\|
v_\theta\!\left(
(1-\tau)\tilde{\mathbf{X}}_{t-1}
+\tau\mathbf{X}^*_t,
\tau;
x^*_{\le t}
\right)
-
\left(
\mathbf{X}^*_t-\tilde{\mathbf{X}}_{t-1}
\right)
\right\|^2.
\end{equation}
This training procedure exposes the model to the source-distribution errors
that arise during recursive deployment, while renoising further introduces robustness by perturbing the source
distributions.

The renoising level $\tau_{r}$ controls the trade-off between preserving information from the previous forecast and robustness to accumulated source errors. At $\tau_{r}=1$, the source reduces to Gaussian noise, recovering the conventional noise-to-target generation paradigm. It eliminates recursive source errors, but meanwhile requires substantially larger sampling steps. At $\tau_{r}=0$, the model directly reuses its raw previous forecast, which may maximize the information encoded in the historical generation but also making subsequent updates most sensitive to errors. We treat $\tau_{r}$ as a hyperparameter to tune in practice. 

Self-rollout and renoising provide complementary roles that are both critical for stable long-horizon deployment: self-rollout exposes the model to the source-distribution errors induced by its own recursive generation, while renoising improves robustness around these model-induced sources. In Section~\ref{sec:error_analysis}, we empirically show that removing either component leads to clear error accumulation over forecasting time.

\textbf{Comparison to self-forcing diffusion.} Although self-forcing~\citep{huang2026self} also
exposes the model to its own generations during training, they consider autoregressive noise-to-target diffusion and feeds generated outputs back to models' conditioning
history, and the diffusion's source distribution remains Gaussian noise. Empirically, we find that this noise-to-target formulation yields suboptimal performance and requires a large number of sampling steps when modeling a long-horizon $\mathbf{X}_t=x_{t+1:t+H}$. In contrast, self-rollout \model unrolls generated forecasts as source samples for
subsequent flow transports, and we find this forecast-to-forecast flow yields better performance-efficiency trade-offs.


\subsection{Stable and Efficient Self-rollout Training}
\label{sec:stable_train}
The self-rollout training directly mitigates the recursive source-distribution mismatch, but naively optimizing it is computationally expensive and unstable. A full rollout over a long forecast episode $T$ requires repeatedly solving the flow ODE, while backpropagating through both the flow ODE steps and the entire rollout chain leads to high memory cost and unstable gradients. We discuss several practical strategies to make self-rollout training stable and efficient. 

Firstly, we apply stop gradient to all model rollouts $\hat{\mathbf{X}}_t$, so that gradients are not propagated through earlier flow ODE and rollout steps. This substantially reduces the backpropagation depth and memory cost. Stop-gradient techniques are also adopted in previous self-forcing training and other flow matching optimization~\citep{huang2026self,geng2025mean}.

Self-rollout also introduces a moving source distribution in flow matching loss, since the source samples $\mathbf{\hat{X}}_{t}$ used for flow matching training are generated by the model itself whose parameters $\theta$ are continuously updated. To reduce the resulting optimization instability, we maintain an exponential moving average of the model parameters, denoted by $\theta_{\mathrm{EMA}}$, and use the EMA model to generate rollout sources. Since $\theta_{\mathrm{EMA}}$ evolves more smoothly than the online parameters $\theta$, the corresponding source distributions vary more slowly during training. The EMA model is also used for inference after training. We find this greatly stabilizes the training and makes optimization easier. 

Prior work also adopts EMA models to stabilize self-distillation training when model-generated
predictions directly define training targets along the generative trajectory
~\citep{song2023consistency,frans2025one}. Self-forcing
~\citep{huang2026self}, by contrast, feeds model generations back as conditioning
context and does not rely on an EMA rollout model. In \model, generated forecasts
directly determine the source distribution of subsequent transports, motivating
EMA to stabilize this evolving source distribution.


Lastly, unrolling over the full trajectory of length $T$ remains computationally expensive even with stopped gradients. Instead, we randomly slice a contiguous trajectory chunk of length $T_0\ll T$ and perform self-rollout only within this $T_0$-step chunk. This truncated rollout substantially reduces training cost while still exposing the model to multi-step error propagation. As shown in Section~\ref{sec:error_analysis}, training with short rollout chunks ($T_0\le 4$) is sufficient for stable deployment over forecasting steps far beyond $T_0$.

With these strategies, self-rollout \model can be trained efficiently and stably. The complete training and inference procedures are described in Algorithms~\ref{alg:sfm_train} and~\ref{alg:sfm_inference}.

\begin{algorithm}[t]
\caption{Self-rollout Seq-Flow Training}
\label{alg:sfm_train}
\begin{algorithmic}[1]

\Require A ground-truth trajectory slice $x^*_{1:T}$ and its corresponding $\mathbf{X}^*_{t}:=x_{t+1:t+H}$, self-rollout step $T_0$, model weights $\theta$ and its EMA weights $\theta_{\text{EMA}}$, EMA parameter $\beta\in [0, 1]$
\State Randomly sample $t_0 \sim \text{Uniform}(1, T-T_0-H+1)$
\For{$t = t_0$ to $t_0+T_0-1$}
\If{$t=t_0$}
    \State $\tilde{\mathbf{X}}_0 \sim \mathcal{N}(0,I)$
\Else
    \State $\tilde{\mathbf{X}}_{t-1}
    =(1-\tau_{r})\hat{\mathbf{X}}_{t-1}
    +\tau_{r}\cdot\mathcal{N}(0,I)$
\EndIf
     \State Sample a random flow interpolation time $\tau\in [0, 1]$
     \State Flow matching loss $\mathcal{L}_t(\theta)=\norm{v_{\theta}((1-\tau)\tilde{\mathbf{X}}_{t-1}+\tau \mathbf{X}^*_t, \tau;x^*_{\le t})-(\mathbf{X}^*_t-\tilde{\mathbf{X}}_{t-1})}^2$
     \State Rollout by solving ODE $\frac{d}{d\tau} \mathbf{Z}_t(\tau)
    =
    v_{\theta_{\text{EMA}}}(\mathbf{Z}_t(\tau), \tau; x_{\le t})$ starting by $\mathbf{Z}_t(0)=\tilde{\mathbf{X}}_{t-1}$, and obtain $\mathbf{Z}_t(1)$.
     \State Collect new forecast $\hat{\mathbf{X}}_t=\text{sg}(\mathbf{Z}_t(1))$
\EndFor
\State \Return updated weights $\theta\leftarrow \theta-\nabla_{\theta}\sum_{t}\mathcal{L}_t(\theta)$ and EMA weights $\theta_{\text{EMA}}\leftarrow (1-\beta)\theta_{\text{EMA}}+\beta\theta.$
\end{algorithmic}
\end{algorithm}

\begin{algorithm}[t]
\caption{Seq-Flow Inference}
\label{alg:sfm_inference}
\begin{algorithmic}[1]
\Require Initial state $x_1$, renoise level $\tau_{r}$, EMA \model model $v_{\theta_{\text{EMA}}}(\mathbf{Z}_t(\tau),\tau;x_{\le t})$, task episode $T$
\State Initialize $\mathbf{Z}_1(0)\sim \mathcal{N}(0,I)$ and solve $\frac{d}{d\tau} \mathbf{Z}_1(\tau)
    =
    v_\theta(\mathbf{Z}_1(\tau), \tau; x_{1})$ to generate the initial forecast $\mathbf{X}_1=\mathbf{Z}_1(1)$
\For{$t = 2$ to $T$}
     \State Receive new observation $x_t$
     \State $\tilde{\mathbf{X}}_{t-1}= (1-\tau_{r})\cdot \mathbf{X}_{t-1}+\tau_{r}\cdot \mathcal{N}(0, I)$
     \State Solve ODE $\frac{d}{d\tau} \mathbf{Z}_t(\tau)
    =
    v_{\theta_{\text{EMA}}}(\mathbf{Z}_t(\tau), \tau; x_{\le t})$ starting from $\mathbf{Z}_t(0)=\tilde{\mathbf{X}}_{t-1}$, and obtain $\mathbf{Z}_t(1)$
     \State Obtain new forecast $\mathbf{X}_t=\mathbf{Z}_t(1)$
\EndFor
\end{algorithmic}
\end{algorithm}

%% file: 4_related.tex
\section{Related Works}
\label{sec:related}
\textbf{Diffusion models for temporal data.}
Diffusion and flow-based models have been increasingly applied to temporal and sequential data, with different designs for incorporating temporal structure into the generative process. One line of work develops asynchronous denoising schedules, where tokens at different physical times are assigned different noise levels~\citep{pmlr-v235-ruhe24a,wu2023ar,chen2024diffusion}. Compared to full-sequence denoising~\citep{li2022diffusion,ho2022video} and autoregressive denoising~\citep{hoogeboom2021autoregressive,rasul2021autoregressive}, asynchronous denoising allows different parts of the sequence to represent different levels of uncertainty during generation. Another line of work incorporates temporal structure into the source distribution itself. Rather than using isotropic Gaussian noise, these methods adopt structured priors, such as Gaussian processes, to introduce temporal correlations that better match the data~\citep{bilovs2023modeling,kollovieh2025flow}. However,  the deployment of these models in online forecasting is still typically re-initialized from a pre-determined source distribution and requires many denoising steps to reach the target distribution. Achieving high-fidelity generation with only one or a few sampling steps is still generally a challenging problem~\citep{liu2023flow,song2023consistency,geng2025mean}.

\textbf{Temporal reuse for efficient diffusion models.}
A growing line of work exploits temporal continuity by reusing previous predictions to accelerate subsequent generation. Warm-start methods perturb predictions from the previous time step and denoise them under the updated observations~\citep{janner2022planning,duan2025real,li2026step}. However, the diffusion model is still trained for noise-to-target generation rather than for the forecast-update transport itself. Asynchronous denoising methods instead distribute denoising computation across physical time by maintaining partially denoised future predictions and progressively updating them as the system evolves~\citep{wei2025cldiffphycon,hoeg2025fast,guo2026efficient}. This pipeline improves sampling efficiency, but the future predictions remain intentionally partially denoised and therefore less accurate at future time steps. Streaming Flow~\citep{jiang2025streaming} and ODEWorld~\citep{liu2026odeworld} are more closely related in aligning flow or ODE dynamics with temporal evolution. Their formulations, however, differ from Seq-Flow in how flow dynamics correspond to physical-time evolution and how new observations are incorporated. We discussed these distinctions in Section~\ref{sec:seqflow}.


\textbf{Error accumulation in recursive generation.}
Prior work studies train--test mismatch from conditioning on ground-truth histories during training but model-generated histories at inference. \model instead faces a flow-matching source-distribution mismatch, as each generated forecast initializes the next flow. We discuss this distinction in Section~\ref{sec:self_rollout}.


%% file: 5_exp.tex
\section{Experiments}
We aim to study the following questions for \model. \textbf{Q1} (\emph{Temporal alignment}): Does aligning the flow ODE with the temporal evolution of forecast distribution improve forecasting in fidelity and efficiency? \textbf{Q2} (\emph{Long-horizon robustness}): Does self-rollout \model control distributional error accumulation over a long run? 

\textbf{Baselines.} We compare \model against several classes of baselines: (1) \emph{Generic diffusion/flow models for sequential data}, including full-trajectory diffusion, autoregressive (AR) diffusion~\citep{rasul2021autoregressive}, and Diffusion Forcing~\citep{chen2024diffusion}. These models all use Gaussian noises as the source distribution for each forecast. We adapt the idea of Diffusion Forcing (training with independently sampled noise levels for each token) to flow model, resulting Flow Forcing. (2) \emph{Diffusion methods with generation reuse}, which exploit previously generated states to facilitate subsequent generation. We implement warm-start diffusion~\citep{janner2022planning, duan2025real}, asynchronous diffusion following~\citet{wei2025cldiffphycon, hoeg2025fast,  guo2026efficient} and StreamingFlow~\citep{jiang2025streaming}; (3) \emph{Self-forcing diffusion}~\citep{huang2026self}, which uses model-generated rollouts during training to reduce the train--test distribution gap in autoregressive diffusion; (4) \emph{Few-step diffusion}: we compare to Meanflow~\citep{geng2025mean}, which is specifically designed to achieve high-quality generation with one or only a few flow sampling steps. 

\textbf{Trajectory alignment.}
Since consecutive $H$-step forecasts are offset by one physical time step, when initializing for the next update,
we drop the first state of the previous forecast and pad its final state,
$
\operatorname{shift}(\mathbf{X}_{t-1})
=
[x_{t+1},\ldots,x_{t+H-1},x_{t+H-1}],
$
so that the overlapping forecast coordinates are aligned in physical time. Other details of the model implementation are deferred to Appendix~\ref{appx:exp_implementation}.

\input{tables/table_main}

\subsection{Scientific Dynamic Systems}

To answer \textbf{Q1}, we first evaluate \model on a synthetic and several scientific forecasting tasks. We primarily consider the beam spill dynamic in the Mu2e experiment at Fermilab, a particle-physics experiment designed to search for new physics~\cite{whitbeck2025fast}.  We further evaluate \model on a simulated and an experimentally-measured real fluid system~\citep{hwang2022solving, hu2026realpdebench}. We present the dataset and task setup below, and the full details is deferred to Appendix~\ref{appx:exp_dataset}.


\textbf{Synthetic system.}
We consider a mixture-of-Gaussian random walk
$
p(x_{t+1}| x_t)
=
\tfrac{1}{2}\mathcal{N}(x_t+1,\delta^2)
+
\tfrac{1}{2}\mathcal{N}(x_t-1,\delta^2),
$
with $\delta=10^{-3}$ and $x_1\sim\mathcal{N}(0,10^{-4})$. 
The task is to infer the $H=5$ future trajectory distribution
$p(x_{t+1:t+5}|x_t)$ over $T=100$ forecast updates. Since the
ground-truth distribution is trackable, this setting
provides a controlled testbed for analyzing distributional fidelity.

\textbf{Beam spill forecasting.}
The Mu2e experiment at Fermilab is a particle-physics experiment designed to
search for charged-lepton flavor violation~\citep{whitbeck2025fast}.
During beam extraction, charged particles are gradually released from the
storage ring, producing a beam intensity that evolves over time. Maintaining
a uniform spill is important for detector live time and for avoiding the loss
of useful physics data~\citep{narayanan2022machine}. The spill-regulation
system operates at 10\,kHz, with a feedback loop on the order of 1\,ms,
making low-latency prediction and continual updating of the evolving beam
state particularly relevant. We use simulated beam-intensity trajectories of
length $430$ and predict the next $H=10$ states conditioned on the observed
history and control inputs.

\textbf{Fluid system forecasting.}
We further evaluate on both simulated and experimentally measured fluid
dynamics. (1) For \textbf{Burgers' equation}, we simulate one-dimensional
velocity fields on a grid of 64 spatial points~\citep{hwang2022solving}.
The model observes only the first 32 points and does not observe the randomly
generated external forcing, making the future evolution uncertain given the
available observations. The task is to forecast future velocities over the
observed half of the domain.
(2) For \textbf{RealPDE-FSI}, we use experimentally measured fluid--structure
interaction data from RealPDEBench~\citep{hu2026realpdebench}. Each state
contains two velocity components, and we downsample the original
$128\times128$ fields to $32\times32$, withholding fine-scale spatial
information and thereby introducing additional forecast uncertainty.




\textbf{Distribution evaluation metrics.}
To evaluate distributional quality of forecasts, we report the Continuous Ranked
Probability Score (CRPS), a proper scoring rule widely used in probabilistic
forecasting~\citep{gneiting2007strictly}. Given a ground-truth future trajectory
$\mathbf{X}_t$ and predictive distribution $\hat{p}_t$, we compute
$
\operatorname{CRPS}(\hat{p}_t,\mathbf{X}_t)
=
\mathbb{E}_{\hat{\mathbf{X}}_t\sim\hat{p}_t}
\|\hat{\mathbf{X}}_t-\mathbf{X}_t\|_1
-
\frac{1}{2}
\mathbb{E}_{\hat{\mathbf{X}}_t,\hat{\mathbf{X}}'_t\sim\hat{p}_t}
\|\hat{\mathbf{X}}_t-\hat{\mathbf{X}}'_t\|_1.
$
The first term measures forecast accuracy, while the second rewards appropriate
predictive dispersion. As a complementary pointwise metric, we also report root
mean squared error (RMSE). For the synthetic task, where the ground-truth
predictive distribution $p_t$ is available, we directly evaluate distributional
fidelity using the average 1-Wasserstein distance
$\frac{1}{T}\sum_{t=1}^{T}\mathcal{W}_1(p_t,\hat{p}_t)$.

\textbf{Shared setup and evaluation.}
We use Transformer backbones for Beam Spill and Burgers, and spatiotemporal U-Nets for FSI. The forecast horizon is $H=5$ for synthetic and $H=10$ for Beam Spill, Burgers, and FSI.

\textbf{Results.} Table~\ref{tab:main_results}  shows that, using three function evaluations per forecast update, \model achieves the lowest $\mathcal{W}_1$ distance among three-NFE methods on the synthetic task ($1.069$, a $29.9\%$ reduction over Flow Forcing), the lowest CRPS and RMSE on Fermilab beam-spill forecasting ($0.015$ and $0.080$, improving over three-NFE full-trajectory flow by $58.3\%$ and $42.4\%$, respectively, while also outperforming its ten-NFE variant), and the lowest CRPS with the lowest or joint-lowest RMSE on Burgers' equation and RealPDE-FSI.

\subsection{Performance-Efficiency Trade-offs}
\label{sec:nfe_tradeoff}
\input{figs/NFE_tradeoff}

To further \textbf{Q1} in terms of performance-efficiency trade-offs, we vary the number of flow sampling steps (NFE) used consistently during both training rollouts and inference for \model, and compare against the other baselines under the same sampling budgets. Figure~\ref{fig:nfe_tradeoff} shows that \model achieves better performance than the Gaussian-source flow matching and other baselines. Particularly, on synthetic task, a 3-NFE \model attains competitive performance to the full-step rectified flow sampling from Gaussian noise. On beam spill forecasting, \model consistently outperform full-step rectified flow for all sampling steps.

\subsection{Ablation Study and Error Accumulation Analysis}
\label{sec:error_analysis}

\input{figs/error_analysis}

\input{figs/effect_T0}

To answer \textbf{Q2}, we study how Seq-Flow's key design choices affect
distributional error over recursive forecast updates. We aim to answer:
(1) whether self-rollout training and renoising are necessary for controlling error
accumulation; (2) whether a model trained with a short self-rollout horizon
$T_0$ can remain stable over substantially longer deployment horizons.

\textbf{Ablation of self-rollout and renoising.}
We compare \model against two ablated variants: one trained with
ground-truth trajectories instead of model-generated self-rollouts, and
one trained with $\tau_{r}=0$. Figure~\ref{fig:error} shows that either
training with ground-truth sources or removing renoising can lead to rapidly accumulating error over
successive forecast updates, consistent with the recursive source-distribution
mismatch described in Section~\ref{sec:error_analysis}. These results show that both self-rollout exposure and source
perturbation are important for robust recursive flow transports.

\textbf{Effect of self-rollout horizon $T_0$.}
We next vary $T_0$ in Algorithm~\ref{alg:sfm_train}, which determines the number
of consecutive forecast updates used during self-rollout training. Figure~\ref{fig:num_rollout} demonstrates that, although the
training rollout step is much shorter than the full task episode, \model remains
stable over substantially longer forecasting runs. In particular, using
$T_0=1$ and $T_0=4$ is sufficient to control error over episodes $T=100$ on synthetic and $T=430$ on Beam Spill respectively, demonstrating that long-horizon
robustness does not require unrolling the full deployment horizon during
training.

%% file: tables/table_main.tex
\begin{table*}[t!] 
\vspace{-12pt} 
    \centering 
    \caption{Results on the synthetic dataset, Beam Spill, Burgers' Equation, and RealPDE-FSI. Prediction horizon $H=5$ for the synthetic task and $10$ for the remaining tasks.} 
    \label{tab:main_results} 
    \resizebox{\textwidth}{!}{ 
    \begin{tabular}{lcccccccc} 
        \toprule 
        \multirow{2}{*}{\textbf{Method}} 
        & \multirow{2}{*}{\textbf{NFE}} 
        & \multicolumn{1}{c}{\textbf{Synthetic}} 
        & \multicolumn{2}{c}{\textbf{Beam Spill}} 
        & \multicolumn{2}{c}{\textbf{Burgers' Equation}} 
        & \multicolumn{2}{c}{\textbf{RealPDE-FSI}} \\ 
        \cmidrule(lr){3-3} 
        \cmidrule(lr){4-5} 
        \cmidrule(lr){6-7} 
        \cmidrule(lr){8-9} 
        & 
        & 
        $\mathcal{W}_1$ dist.$\downarrow$ 
        & CRPS $\downarrow$ 
        & RMSE $\downarrow$ 
        & CRPS $\downarrow$ 
        & RMSE $\downarrow$ 
        & CRPS $\downarrow$ 
        & RMSE $\downarrow$ \\ 
        \midrule 
 
        AR Diffusion 
        & $H \times 3$ 
        & $1.270$ 
        & $0.363$ & $1.001$ 
        & $0.0589$ & $0.1183$ 
        & $0.0150$ & $0.0260$ \\ 
 
        Self-forced AR Diffusion 
        & $H \times 3$ 
        & $2.765$ 
        & $0.055$ & $0.209$ 
        & $0.0143$ & $0.0395$ 
        & $0.0031$ & $0.0062$ \\ 
 
        Full-trajectory Diffusion 
        & 3 
        & $2.260$ 
        & $0.043$ & $0.179$ 
        & $0.0139$ & $0.0366$ 
        & $0.0030$ & $0.0063$ \\ 
 
        AR Flow 
        & $H \times 3$ 
        & $1.203$ 
        & $0.446$ & $1.092$ 
        & $0.0822$ & $0.1401$ 
        & $0.0130$ & $0.0208$ \\ 
 
        Self-forced AR Flow 
        & $H \times 3$ 
        & $2.872$ 
        & $0.059$ & $0.230$  
        & $0.0198$ & $0.0356$ 
        & $0.0039$ & $0.0081$ \\ 
 
        Full-trajectory Flow 
        & 3 
        & $2.221$ 
        & $0.036$ & $0.139$ 
        & $0.0068$ & $0.0175$ 
        & $0.0029$ & $0.0056$ \\ 
 
        Full-trajectory Flow 
        & 10 
        & $\textbf{0.905}$ 
        & $0.035$ & $0.151$ 
        & $0.0066$ & $0.0178$ 
        & $0.0028$ & $0.0062$ \\ 
 
        Diffusion Forcing 
        & 3 
        & $1.533$ 
        & $0.050$ & $0.240$ 
        & $0.0137$ & $0.0357$ 
        & $0.0040$ & $0.0080$ \\ 
 
        Flow Forcing 
        & 3 
        & $1.526$ 
        & $0.040$ & $0.167$ 
        & $0.0100$ & $0.0205$ 
        & $0.0030$ & $\textbf{0.0055}$ \\ 
 
        MeanFlow 
        & 3 
        & $50.081$ 
        & $0.103$ & $0.631$ 
        & $0.7253$ & $2.7046$ 
        & $0.0028$ & $0.0059$ \\ 
 
        StreamingFlow 
        & $H \times 1$ 
        & $2.381$ 
        & $0.469$ & $2.485$ 
        & $0.0706$ & $0.3114$ 
        & $0.0029$ & $0.0083$ \\ 
 
        Warm-start Flow 
        & 3 
        & $304.403$ 
        & $1.349$ & $6.051$ 
        & $0.0191$ & $0.0391$ 
        & $0.0027$ & $0.0059$ \\ 
 
        Asynchronous Flow 
        & 3 
        & $20.766$ 
        & $0.522$ & $0.912$ 
        & $0.0572$ & $0.0935$ 
        & $0.0067$ & $0.0202$ \\ 
 
        \midrule 
 
        \textbf{\model} 
        & 3 
        & $1.069$ 
        & $\textbf{0.015}$ & $\textbf{0.080}$ 
        & $\textbf{0.0060}$ & $\textbf{0.0165}$ 
        & $\textbf{0.0026}$ & $\textbf{0.0055}$ \\ 
 
        \bottomrule 
    \end{tabular} 
    } 
    \vspace{-6pt} 
\end{table*}

%% file: figs/NFE_tradeoff.tex
\begin{figure*}[t]
    \centering
    \includegraphics[width=0.48\textwidth]{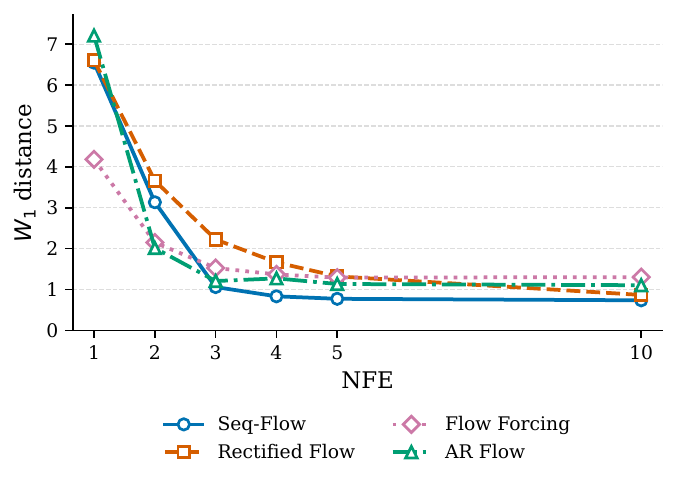}
    \includegraphics[width=0.48\textwidth]{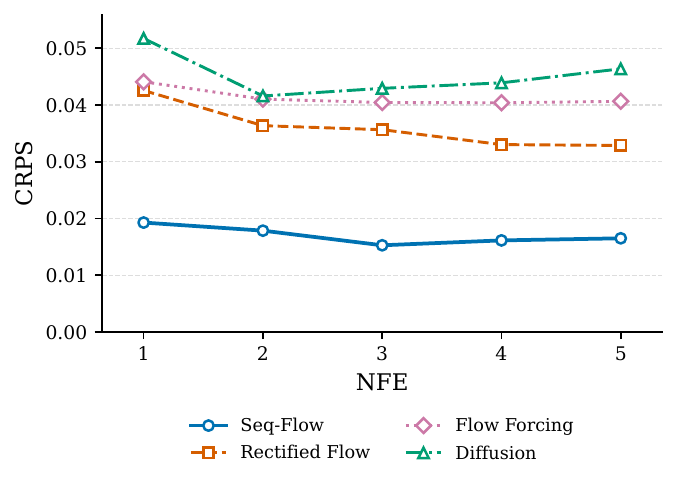}
    \caption{\textbf{Performance-efficiency trade-offs}. X axis: sampling steps (NFE), Y axis: task performance. For clarity, each figure shows only a subset of the top-performing baselines. Left: synthetic task. Right: beam spill forecasting.}
    \vspace{-7pt}
    \label{fig:nfe_tradeoff}
    \vspace{-6pt}
\end{figure*}

%% file: figs/error_analysis.tex
\begin{figure*}[t]
    \centering
    \includegraphics[width=0.48\textwidth]{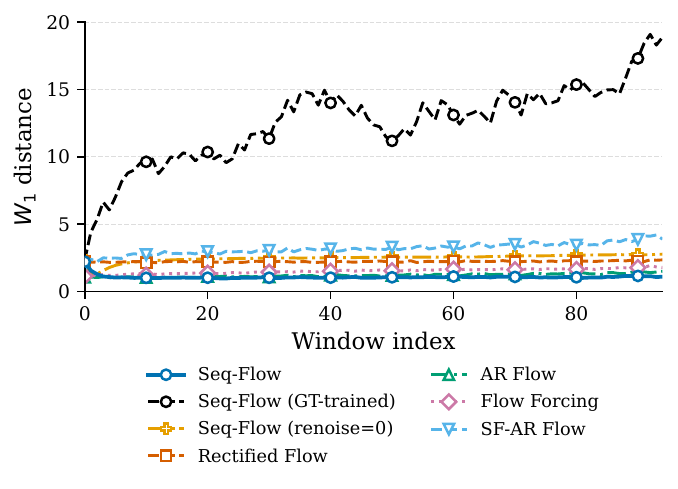}
    \includegraphics[width=0.48\textwidth]{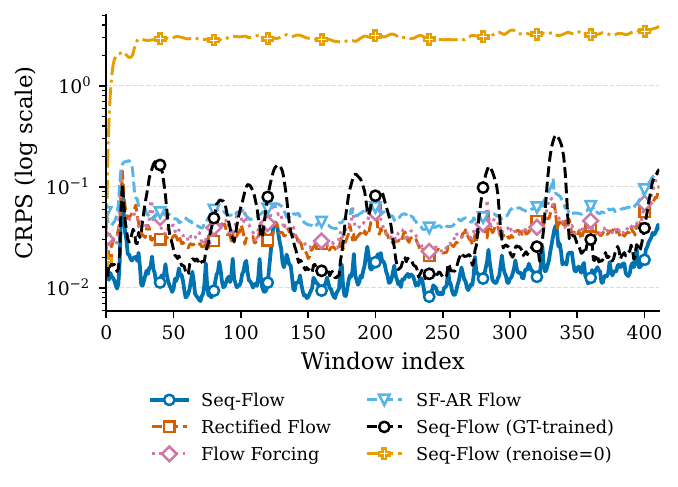}
    \vspace{-4pt}
    \caption{\textbf{\model with controlled error accumulation over forecast steps}.  Each figure shows performance over forecast step (window id). Left: synthetic task. Right: beam spill forecasting.}
    \label{fig:error} 
    \vspace{-5pt}
\end{figure*}

%% file: figs/effect_T0.tex
\begin{figure*}[t]
    \centering
    \includegraphics[width=0.48\textwidth]{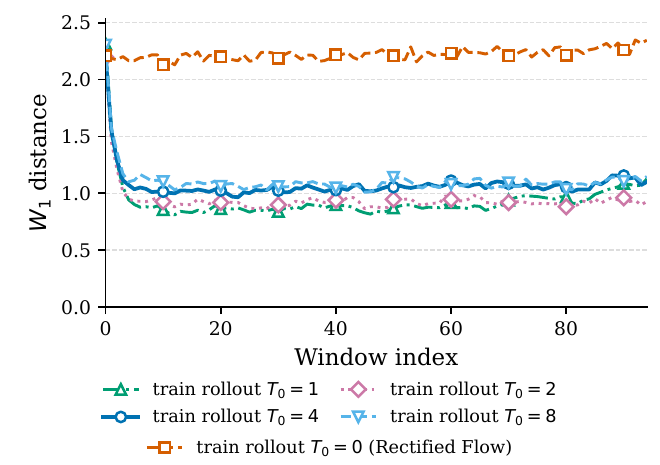}
    \includegraphics[width=0.48\textwidth]{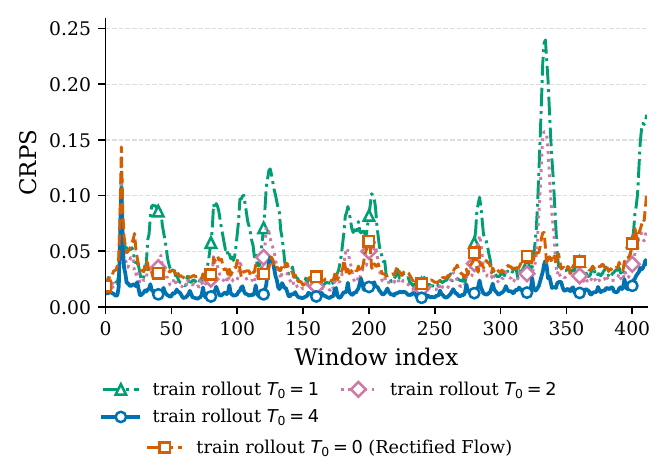}
    \vspace{-4pt}
    \caption{\textbf{Short self-rollout steps $T_0$ leads to robust long-run forecasting}. Each figure shows performance per forecast step with different $T_0$. Left: synthetic task. Right: beam spill forecasting.}
    \label{fig:num_rollout} 
\end{figure*}

%% file: 6_appendix.tex
%
%

\clearpage
\section{Experimental Details}
\label{appx:exp}

\subsection{Datasets and Preprocessing}
\label{appx:exp_dataset}

Table~\ref{tab:appendix_datasets} summarizes the data. Training uses contiguous clips; evaluation updates the conditioning history with observed, not generated, states.

\paragraph{Synthetic random walk.}
We use a scalar random walk with initial state $x_1\sim\mathcal{N}(0,10^{-4})$ and transition law
\[
    p(x_{t+1}\mid x_t)
    = \tfrac12\mathcal{N}(x_{t+1};x_t+1,\delta^2)
    + \tfrac12\mathcal{N}(x_{t+1};x_t-1,\delta^2),
    \qquad \delta=10^{-3}.
\]
Each increment independently chooses $+1$ or $-1$ with equal probability and adds Gaussian noise. No normalization is applied.

The task is to forecast the next five states given the observed history. Since the process is Markov, the exact forecast depends only on the latest state $x_t$. For lead $h=1,\ldots,5$, its marginal is
\[
    p(x_{t+h}\mid x_t)
    = \sum_{k=0}^{h}2^{-h}\binom{h}{k}
      \mathcal{N}(x_{t+h};x_t+2k-h,h\delta^2),
\]
where $k$ counts positive increments. These analytic marginals provide a ground-truth reference for evaluating predictive distributions.

\paragraph{Beam Spill.}
The Mu2e simulator~\citep{narayanan2022machine} produces trajectories of scalar beam intensity and control signals. Samples are recorded every $0.1\,\mathrm{ms}$, with 430 samples covering a $43\,\mathrm{ms}$ spill. Temporally correlated disturbances make future intensity uncertain given the observed history.

Our task is conditional forecasting, not controller learning. At each update, models receive observed intensities and the supplied controls over the forecast interval, but not future intensities. They predict the next 10 intensity values, corresponding to $1\,\mathrm{ms}$. We train on 40,000 trajectories and evaluate on 5,000 held-out trajectories, using the supplied intensity and control normalization statistics.

\paragraph{Burgers' equation.}
Following~\citet{wei2025cldiffphycon}, we simulate a scalar velocity field $u(x,t)$ governed by the forced one-dimensional Burgers' equation on $x\in[0,1]$ and $t\in[0,1.98]$:
\[
    \begin{aligned}
        \partial_t u &= -u\,\partial_xu+\nu\,\partial_{xx}u+f,
        \qquad \nu=0.01,\\
        u(0,t)&=u(1,t)=0,\qquad u(x,0)=u_0(x).
    \end{aligned}
\]
Here $u_0$ is the initial velocity field and $f$ is external forcing. We use 64 interior spatial points and an integration step of $10^{-4}$, recording states every $0.02$ time units. Initial fields combine two Gaussian profiles with random locations, widths, and amplitudes; forcing combines eight localized space--time Gaussians.

Models observe only the first 32 spatial points and forecast the next 10 velocity fields on that same observed region. Velocities are scaled by $1/10$. The remaining spatial points and the forcing are withheld from the model. Thus, although the full dynamics are deterministic, the observed history does not specify the hidden state and forcing, making the forecast uncertain.

\paragraph{RealPDE-FSI.}
We use measured $(u,v)$ velocity fields from the fluid--structure interaction dataset in RealPDEBench~\citep{hu2026realpdebench}. Each state contains two velocity components on a spatial grid. We retain every fourth point of the original $128\times128$ grid, giving two $32\times32$ channels without changing the temporal sampling. This coarsening withholds fine-scale spatial information from the model.

The task is to forecast the next 10 velocity fields from the observed history, with both velocity components included in the prediction. Channels are standardized using training statistics. We train on 39 trajectories and evaluate on six held-out trajectories.

\begin{table}[H]
    \centering
    \caption{Dataset settings. $L$: episode length; $H$: forecast horizon. Counts are trajectories; field shapes are channels $\times$ grid.}
    \label{tab:appendix_datasets}
    \footnotesize
    \begin{tabular*}{\linewidth}{@{\extracolsep{\fill}}llllrr@{}}
        \toprule
        Dataset & State shape & Training & Held-out & $L$ & $H$ \\
        \midrule
        Synthetic & $1$ & 50,000 & 5,000 & 100 & 5 \\
        Beam Spill & $1$ & 40,000 & 5,000 & 430 & 10 \\
        Burgers & $32$ & 90,000 & 10,000 & 100 & 10 \\
        RealPDE-FSI & $2\times32\times32$ & 39 & 6 & 100 & 10 \\
        \bottomrule
    \end{tabular*}
\end{table}

\subsection{Model and Optimization Settings}
\label{appx:exp_implementation}

\paragraph{Backbones.}
Synthetic and Beam Spill have scalar states, while Burgers has 32-dimensional observed states; these three tasks use a Transformer backbone. RealPDE-FSI uses a spatiotemporal U-Net for its two-channel spatial velocity fields. Table~\ref{tab:appendix_backbones} groups the settings by architecture. Input and output shapes follow Table~\ref{tab:appendix_datasets}.

\begin{table}[H]
    \centering
    \caption{Backbone configurations. U-Net channel multipliers scale the base width.}
    \label{tab:appendix_backbones}
    \small
    \begin{tabular*}{\linewidth}{@{\extracolsep{\fill}}lr@{\hspace{2em}}lr@{}}
        \toprule
        \multicolumn{2}{c}{\textbf{Transformer}} & \multicolumn{2}{c}{\textbf{Spatiotemporal U-Net}} \\
        \multicolumn{2}{c}{Synthetic, Beam Spill, Burgers} & \multicolumn{2}{c}{RealPDE-FSI} \\
        \cmidrule(lr){1-2}\cmidrule(lr){3-4}
        Hidden width & 128 & Base width & 64 \\
        Layers & 12 & Channel multipliers & $(1,2,4)$ \\
        Feed-forward width & 512 & Attention heads & 4 \\
        Attention heads & 4 & Head dimension & 32 \\
        \bottomrule
    \end{tabular*}
\end{table}

\paragraph{Optimization and training.}
All four Seq-Flow configurations use AdamW with weight decay $10^{-4}$ and $(\beta_1,\beta_2)=(0.9,0.999)$. Table~\ref{tab:appendix_seqflow_settings} reports the dataset-specific learning rates, batch sizes, warmup steps, and checkpoint steps. We use an exponential moving average (EMA) with decay 0.999. Training combines Gaussian-source and self-rollout losses with equal weights of 0.5, and each training rollout includes four forecast updates.

\begin{table}[H]
    \centering
    \caption{Seq-Flow settings. $C_0/C_{\max}$: initial/maximum context length. Ablations vary NFE or rollout length.}
    \label{tab:appendix_seqflow_settings}
    \footnotesize
    \begin{tabular*}{\linewidth}{@{\extracolsep{\fill}}lcccc@{}}
        \toprule
        Setting & Synthetic & Beam Spill & Burgers & RealPDE-FSI \\
        \midrule
        Learning rate & $5\times10^{-4}$ & $5\times10^{-4}$ & $5\times10^{-4}$ & $2\times10^{-4}$ \\
        Effective batch size & 4096 & 512 & 64 & 32 \\
        Weight decay & $10^{-4}$ & $10^{-4}$ & $10^{-4}$ & $10^{-4}$ \\
        AdamW $(\beta_1,\beta_2)$ & $(0.9,0.999)$ & $(0.9,0.999)$ & $(0.9,0.999)$ & $(0.9,0.999)$ \\
        Warmup steps & 10,000 & 10,000 & 10,000 & 1,000 \\
        Checkpoint step & 12k & 36k & 36k & 10k \\
        Initial/update NFE & $3/3$ & $3/3$ & $3/3$ & $5/3$ \\
        Renoising level & 0.4 & 0.4 & 0.4 & 0.4 \\
        $C_0/C_{\max}$ & $1/2$ & $9/10$ & $9/10$ & $9/10$ \\
        Forecast windows per episode & 95 & 412 & 82 & 82 \\
        \bottomrule
    \end{tabular*}
\end{table}

\paragraph{Sequential evaluation.}
Initial/update NFE in Table~\ref{tab:appendix_seqflow_settings} denotes the number of function evaluations for the first forecast and subsequent forecast updates, respectively. The context lengths $C_0$ and $C_{\max}$ specify the initial and maximum observed history. The four-update training rollout does not limit the evaluation episode: each episode has $L-C_0-H+1$ forecast windows, including initialization.